\documentclass[11pt,a4paper]{article}
\usepackage[table,xcdraw]{xcolor} 
\usepackage[hyperref]{acl2020}
\usepackage{times}
\usepackage{latexsym}

\usepackage{graphicx}
\graphicspath{ {images/} }
\usepackage[autostyle]{csquotes}
\usepackage{amsmath,amssymb,amsfonts, mathtools}
\usepackage{placeins}

\usepackage{microtype}

\aclfinalcopy 
\def\aclpaperid{3207} 

\title{Location Attention for Extrapolation to Longer Sequences}

\author{Yann Dubois \\
  University of Cambridge \\
  \texttt{yanndubois96@gmail.com} \\ \And 
  Gautier Dagan \\
  University of Amsterdam \\
  \texttt{gautier.dagan@gmail.com} \\\AND
  Dieuwke Hupkes\thanks{\ \ Shared senior authorship} \\
  ILLC, University of Amsterdam \\
  \texttt{d.hupkes@uva.nl} \\\And
   Elia Bruni\textsuperscript{*} \\
  Universitat Pompeu Fabra \\
  \texttt{elia.bruni@gmail.com}
  }

\date{}

\begin{document}
\maketitle

\begin{abstract}
Neural networks are surprisingly good at interpolating and perform remarkably well when the training set examples resemble those in the test set.
However, they are often unable to extrapolate patterns beyond the seen data, even when the abstractions required for such patterns are simple.
In this paper, we first review the notion of extrapolation, why it is important, and how one could hope to tackle it. 
We then focus on a specific type of extrapolation, which is especially useful for natural language processing: generalization to sequences longer than those seen during training.
We hypothesize that models with a separate content- and location-based attention are more likely to extrapolate than those with common attention mechanisms.
We empirically support our claim for recurrent \textit{seq2seq} models with our proposed attention on variants of the \textit{Lookup Table} task.
This sheds light on some striking failures of neural models for sequences and on possible methods to approaching such issues.

\end{abstract}

\section{Introduction}

It is indisputable that, in recent years, neural network research has made stunning progress on a wide variety of tasks that require to process sequential inputs, such as machine translation \citep{sutskever2014sequence} and speech recognition \citep{graves2013speech}. 
However, many researchers have questioned the forms of generalization that neural networks exhibit, which significantly diverges from human-like generalization \citep{baroni2018still,geirhos2018generalisation}.
This discrepancy with human-like generalization is particularly true when it comes to extrapolating ``outside'' the training space \citep{delosh1997extrapolation,marcus1998rethinking}.

As neural networks are powerful memorizers  \cite{zhang2016understanding} and easily learn superficial statistical cues \cite{jo2017measuring}, testing extrapolation and generalization to samples from the long tails of a distribution 
might be the only way of quantifying their capacity of abstract reasoning \cite{barrett2018measuring}. 

Despite this benefit, little work has been done in extrapolation. A possible explanation is that the probability of encountering a test example in the extrapolation setting seems low when the training set $\mathcal{D}$ is large.\footnote{Extrapolation is still prevalent in practical scenarios as high-dimensional problems would typically require an exponentially large $\mathcal{D}$ to be representative, and the underlying distribution may vary over time \cite{hooker2004diagnostics}.} However, such an argument fails to consider the high cost of error in extrapolation settings, and this can be a barrier for real-world scenarios (e.g., self-driving cars).

In this paper, we focus on extrapolation in sequences. More precisely, how to generalize sequence-to-sequence predictors to inputs of length $n_{*}>n_{\mathcal{D}}$, where $n_{\mathcal{D}}$ denotes the length of the longest sequence in the training set.
Such extrapolation is crucial for language acquisition, where humans have limited learning resources to account for the unbounded nature of language. 
To successfully generalize, a language learner needs to process new and potentially longer sentences than previously encountered ones \cite{chomsky1956three}. 

Accounting for this unbounded nature of language is challenging for neural networks.
This issue has recently been uncovered for seq2seq models by looking at simple artificial tasks \citep{lake2017generalization,liska2018memorize,weber2018fine}.
\citet{liska2018memorize} find that seq2seq architectures can converge to local minima that generalize, but rarely do. 
This suggests that neural networks could generalize but lack inductive biases that favor extrapolatable behavior.

In the following sections, we review the concepts of attention and extrapolation.
We then argue that current attention mechanisms, which are mainly responsible for recent successes in natural language processing (NLP), are unlikely to extrapolate as they depend on the \emph{content} of trained embeddings.
This leads us to introduce a novel \emph{location}-based attention that is loosely inspired by human visual attention. 
To avoid gaining extrapolation capabilities at the cost of expressivity, we introduce an \emph{attention mixer} that combines content- and position-based attentions. 
Finally, we show that recurrent models equipped with this new attention mechanism can extrapolate to longer sequences.


\section{Extrapolation}

Extrapolation is often used but rarely formally defined.
\citet{ebert2014interpolation} have found that when extrapolation is explicitly defined, it often refers to points outside a hull delimited by the training set.
E.g., rectangular hull, concave hull, or convex hull.
In this work we use the rectangle hull definition \cite{brooks1988characterizing}, as any model which is extrapolatable for this region would also be extrapolatable for the convex and concave definition.

\begin{figure}[h]
\centering
\includegraphics[width=\columnwidth]{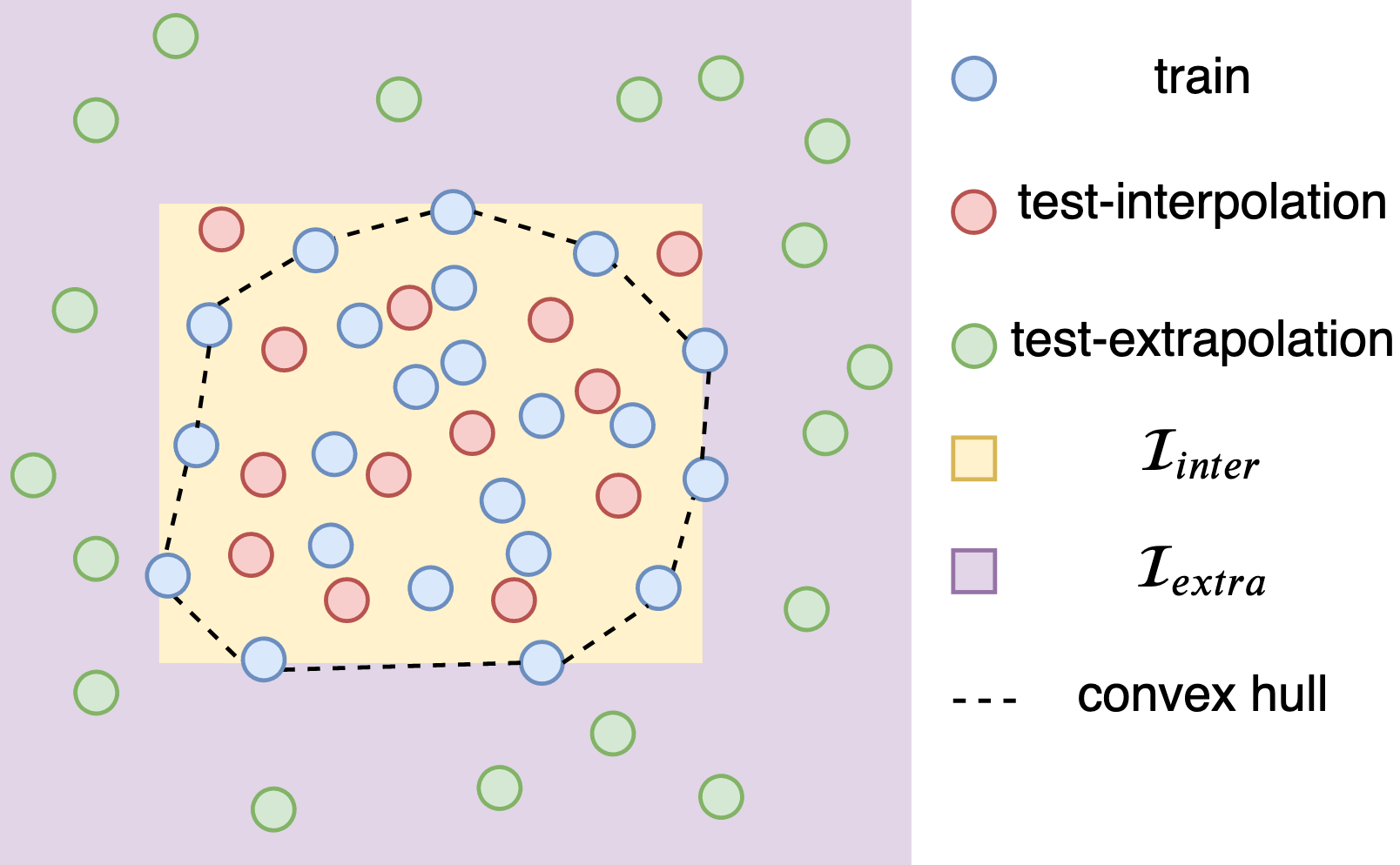}
\caption{Schematic extrapolation setting for $d=2$.}
\label{fig:convhull}
\end{figure}

Given any finite training dataset $\mathcal{D} \coloneqq \{\mathbf{x}^{(n)}\}_{n=1}^N \subset \mathbb{R}^d$, we define the interpolation domain to be the $d$-dimensional interval $\mathcal{I}_{inter} \coloneqq \prod_{i=1}^d [\min_n x_i^{(n)}, \max_n x_i^{(n)}]$ and the \textbf{extrapolation domain} its complement $\mathcal{I}_{extra} \coloneqq \mathbb{R}^d \setminus \mathcal{I}_{inter}$. 
In other words, we define a test example $\mathbf{x}^*$ to be in the extrapolation setting if \textit{at least} one of its features $x^*_j$ is larger or smaller than any values it took during training (Figure~\ref{fig:convhull}).

Throughout this paper, we assume that neural networks with inputs or temporary representations in $\mathcal{I}_{extra}$ will break.
Indeed, for a given target function $t : \mathbb{R}^d \to \mathbb{R}$ to approximate, there is an infinite amount of predictors that satisfy $f(\mathbf{x}) = t(\mathbf{x}),\ \forall \mathbf{x} \in \mathcal{I}_{inter} \subset \mathbb{R}^d$.
Without any additional constraints, it is thus extremely unlikely that $f(\mathbf{x}) = t(\mathbf{x}),\ \forall \mathbf{x} \in \mathbb{R}^d$.
This could explain why neural networks have empirically been found to break in extrapolation settings \cite{lohninger1999teach,hettiarachchi2005extrapolation,mitchell2018extrapolation}. 

The rest of the paper discusses how to constrain representations used by our neural models \footnote{Although the sentence length is a scalar, the temporary representations (outputs of a hidden layer) are high dimensional.} to be in $\mathcal{I}_{inter}$ regardless of the source sentence length, without decreasing their expressivity.


\label{sec:extrap_def}

\section{Desiderata}

First and foremost, we would like a model that can extrapolate to sequences longer than the longest training one $n_{\mathcal{D}}$ (\emph{Extrapolation Constraint}). 
As previously discussed, models with inputs or temporary representations in $\mathcal{I}_{extra}$ will very likely break. 
To satisfy the extrapolation constraint, neural models should thus not depend on features that take values in $\mathcal{I}_{extra}$ for sequences longer than $n_{\mathcal{D}}$.

Second, our model should be able to learn very complex positional attention patterns (\emph{Positional Patterns Constraint}).
Finally, although the position of words in a sentence is important, many tasks depend on their semantics. 
The model should thus still be able to learn content-based attention patterns  (\emph{Content Patterns Constraint}).

In the following section, we review previously proposed attention-mechanism and discuss why they do not fulfill the three aforementioned desired properties.



\label{sec:desiderata}

\section{Attention Mechanisms}


An attention mechanism (or attender) takes as input a matrix of keys $\mathrm{K} \coloneqq \{\mathbf{k}_s^T\}_{s=1}^{n_s} \in \mathbb{R}^{n_s \times d}$ and a query $\mathbf{q}_t \in \mathbb{R}^d$, and outputs a probability mass function $\pmb{\alpha}_t \in \mathbb{R}^{n_s}$ that will weight a set of values $\mathrm{V} \coloneqq \{\mathbf{v}_s^T\}_{s=1}^{n_s} \in \mathbb{R}^{n_s \times d_v}$ to generate a glimpse vector $ \mathbf{g}_t \in \mathbb{R}^{d_v}$ used for downstream tasks. 
Following \citet{graves2014neural}, it is useful to think of the attender as a memory access module, $\pmb{\alpha}_t$  as the soft address and $ \mathbf{g}_t $ as the accessed vector.
\begin{equation}\label{eq:attender}
\mathbf{g}_t \coloneqq \sum_{s=1}^{n_s}   \mathbf{v}_s \mathrm{attender}(\mathbf{k}_s,\mathbf{q}_t) = \mathrm{V} \pmb{\alpha}_t
\end{equation}

\begin{figure}[h!]
\centering
\includegraphics[width=\columnwidth]{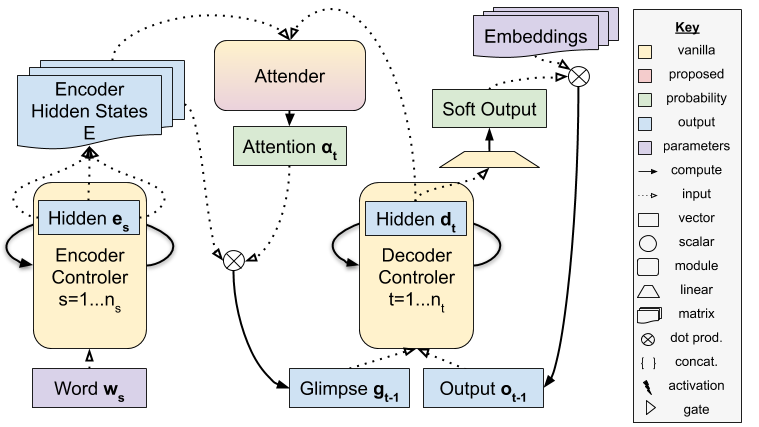}
\caption{Attender in a recurrent seq2seq.}
\label{fig:attender_seq2seq}
\end{figure}

Figure~\ref{fig:attender_seq2seq} illustrates attention in a recurrent seq2seq \cite{cho2014learning}, which we will use for our experiments. 
Both the keys and the values correspond to the set of encoder hidden states $\mathrm{K}=\mathrm{V} = \mathrm{E} \coloneqq \{\mathbf{e}^T_s\}_{s=1}^{n_s}$, while the query corresponds to the current decoder hidden state $\mathbf{q}_t = \mathbf{d}_t$.

\subsection{Content Attention}

Most attention mechanisms compute ``content-based addressing'' (associative memory) that depend on (partial) matches of the key and query.
They take as input $\mathrm{K}$ and $\mathbf{q}_t$ and output a semantic-based attention $\pmb{\gamma}_t \in \mathbb{R}^{n_s}$. 
For example, if you wanted to translate a scientific paper, you could understand the main point of the text without remembering the specific technical terms that were used. 
When translating, you would go back to the text and translate the jargon by knowing what to look for. 

A number of content-based have been proposed, they usually differ in a $\mathrm{score}$ that quantifies the match between $\mathbf{k}_s,\mathbf{q}_t$ through multinomial logits:
\begin{gather}\label{eq:gamma}
\pmb\gamma_{t} \coloneqq \{\mathrm{softmax}(\mathrm{score}(\mathbf{k}_s,\mathbf{q}_t ))\}_{s=0}^{n_s-1} \\[10pt]
\begin{aligned}\label{eq:content}
&\mathrm{score}(\mathbf{k}_s,\mathbf{q}_t ) \coloneqq \\[10pt]
&\begin{cases}
\mathbf{u}^T \tanh([\widetilde{\mathbf{k}}_s ; \widetilde{\mathbf{q}}_t]) & \text{\small \textit{Additive} } \text{\scriptsize \citet{bahdanau2014neural}} \\[5pt]
\mathbf{k}_s^T \widetilde{\mathbf{q}}_t & \text{\small \textit{Multiplicative} } \text{\scriptsize \citet{luong2015effective}} \\[5pt]
\frac{\mathbf{k}_s^T \mathbf{q}_t}{\sqrt{d}} & \text{\small \textit{S. Dot Prod.} } \text{\scriptsize \citet{vaswani2017attention}}   \\
\end{cases}
\end{aligned}
\end{gather}
Where $\widetilde{\mathbf{x}}$ is a shorthand for $W\mathbf{x}$.
\subsection{Location Attention}
A location (or position) attention mechanism computes ``location-based addressing'' (random access memory) that depend on the index of the key. 
It takes as input $\mathbf{q}_t$ and outputs a location attention $\pmb{\lambda}_t \in \mathbb{R}^{n_s}$.
Intuitively, it decides which value to retrieve based on its index. 
For example, in German sentences, the verb goes at the end of the sentence, after a subordinate clause.
When translating from German to English, it might thus make sense to directly attend to the last word in the German source sentence after encoding a subordinate clause.
There are many other cases where attending to words based on their positions seems important. E.g. translating from subject-object-verb to subject-verb-object languages, or understanding the emphasis in some languages.

Despite the importance of word ordering in natural language, location-based attention is not common in \textit{seq2seq} frameworks. 
This is probably because content-based attention can emulate location-based attention in the usual interpolation setting.
Indeed, it can learn to encode a positional embedding in the hidden states of the encoder through some internal ``counter''.
This counter is unlikely to work in the extrapolation regime,\footnote{This assumption can depend on the architecture and the inductive bias it provides \cite{weiss2018practical}. For our task, we found that the assumption held for both LSTM and GRU. } we, therefore, investigate other types of location-attention that could satisfy the extrapolation constraint.

\citet{luong2015effective} proposed a location-based attention by using Equation \ref{eq:gamma} with a score that is independent of the key $\mathrm{score}(\mathbf{k}_s,\mathbf{q}_t )=\mathbf{w}^T \mathbf{q}_t$.
They restrict themselves to sequences of the same length, which is not of interest to our work. 
Such a mechanism could be extended to sequences of varying lengths but would still lack extrapolation capacity as the model still has to learn to embed the location of the index it wants to retrieve.

The Neural Turing Machine
\cite{graves2014neural}, post-processes the content attention by shifting its location by a predicted number of steps.
We use a similar mechanism, which is extrapolatable due to the independence of the sequence length.
Nevertheless, on its own, it does not allow positional-only patterns in variable-length sentences. 
For example, it cannot attend to the i\textsuperscript{th} word irrespective of the sentence length. 
The same argument holds for other location-based attention developed for architectures with an external memory \cite{sukhbaatar2015end}.

More recently, many location-based attention have been proposed in self-attention mechanism. 
These methods are usually based on sinusoidal encodings (SE), which have been proposed to take into account the word positions while bypassing the need for recurrences in encoder-decoder frameworks.
In this paper, we will consider the \textit{transformer} and \textit{transformerXL} (relative SE) attention, which are computed as follows.
\begin{equation}\label{eq:location}
\setlength{\jot}{10pt}
\begin{aligned}
&\mathrm{score}(\mathbf{k}_s,\mathbf{q}_t ) \coloneqq \\
&\begin{cases}
\frac{(\mathbf{k}_s + \mathbf{p}_s)^T (\mathbf{q}_t + \mathbf{p}_t)}{\sqrt{d}} & \text{\small \textit{Transformer} \cite{vaswani2017attention}}   \\[10pt]
\frac{(\widetilde{\mathbf{k}}_s + \widetilde{\mathbf{p}}_{s-t})^T (\widetilde{\mathbf{q}}_t + b)}{\sqrt{d}}  & \text{\small\textit{TransformerXL} \cite{dai2019transformer}}  \\
\end{cases}
\end{aligned}
\end{equation}

Where $\mathbf{p}_{t}$ is a positional encoding with sinusoidals of different frequencies at every dimension.
Although powerful, the sinusoidal encoding and its variants \cite{shaw2018self,dai2019transformer} lack the ability to model location patterns that depend on general word position such as ``look at the $i^{th}$ word (after ...)'' in the extrapolation setting.
Indeed, the sinusoidal encoding for any fixed offset $\mathbf{p}_{t+k}$ is linear in $\mathbf{p}_{t}$ but not in $k$.

Location-based processing of attention has also been proposed as a way of constraining content-based attention to some (soft) window. \citet{yang2018modeling} achieve it by multiplying the content attention by the weights of a predicted Gaussian such that the model has an inductive bias towards attending to words that are close to each other.
\citet{sukhbaatar2019adaptive} use a piece-wise window to decrease the computational complexity of the model.
These methods nevertheless solve a fundamentally different problem and do not allow location-only extrapolatable patterns of attention.

\label{sec:back_attn}

\section{Model}
In this section, we propose a location attender that can satisfy the extrapolation and positional patterns constraint. 
We then discuss how to incorporate content attention to satisfy the content patterns constraint.
%

\subsection{Location Attender}
\label{subsec:location_attender}

We would like our position attention to be loosely reminiscent of human attention, whereby we sequentially focus on a single area of the input (e.g., words or pixels) but vaguely perceive neighboring inputs due to the eccentricity effect \cite{carrasco1995eccentricity}. 
The visual acuity of humans is uni-modal, symmetric, and spikes at the fovea, which corresponds to a $0^\circ$ retinal eccentricity. 
We model this visual acuity using a Gaussian Probability Density Function (PDF) similarly to \citet{mnih2014recurrent}.\footnote{Visual acuity is distributed in a Laplace-like distribution, but initial experiments were more encouraging using a Gaussian.}
I.e. for each step, the Location Attender models a Gaussian attention over the relative word positions.

Specifically, it generates a mean $\mu_t$ and standard deviation $\sigma_t$, which are used to compute the location attention given the values of the PDFs at the relative indices $r_s \coloneqq \frac{s}{n_s-1}$ of the keys:
$$\pmb{\lambda}_{t} \coloneqq \{\mathrm{PDF}_{\mu_t, \sigma_t}(r_s)\}_{s=0}^{n_s-1}$$

Using relative indices $r_s$ instead of the absolute ones $s$ is crucial such that the generated $\mu_t$ is bounded (in $ [0,1]$), thereby satisfying the extrapolation constraint.

This model, unfortunately, fails to satisfy the positional patterns constraint, as it only allows patterns of attention based on percentile positions. E.g., it can decide to attend to the 10\%-percentile word but not to the $2^{nd}$ word.
This incapacity to satisfy the position pattern constraint is a general issue with commonly used attention mechanisms (including sinusoidal-based) that only becomes apparent when dealing with complex extrapolation patterns.

To have a general attention mechanism, we need a $\mu_t$ that can: i) attend to locations based on absolute positions; ii) attend to locations based on percentile positions;
iii) attend to positions based on the previous attention.
We achieve this by defining one building block for each of those requirements ($\mathbf{b}_t$) such that their weighted average forms $\mu_t$, and the weights $\pmb{\rho}_t$ are bounded outputs of the model. The three building blocks are:

\begin{itemize}
\item The step size $\frac{1}{n_s-1}$ between words allows the attention mechanism to depend on absolute positions. The generated weight is an integer, which dictates the additional number of steps to take.
\item The bias term $1$ enables the model to use percentile positions. The generated weight gates it (on or off). 
\item The average position of the previous attention $\bar{\pmb{\alpha}}_{t-1}$ that is gated by the generated weight. This ensures that the model can attend using absolute positions to words at indices not seen during training. E.g., attending to index $n_\mathcal{D}+5$ by first attending to $n_\mathcal{D}$ then $\bar{\pmb{\alpha}}_{t-1} + 5$.
\end{itemize}

\begin{figure}[h]
\centering
\includegraphics[width=\columnwidth]{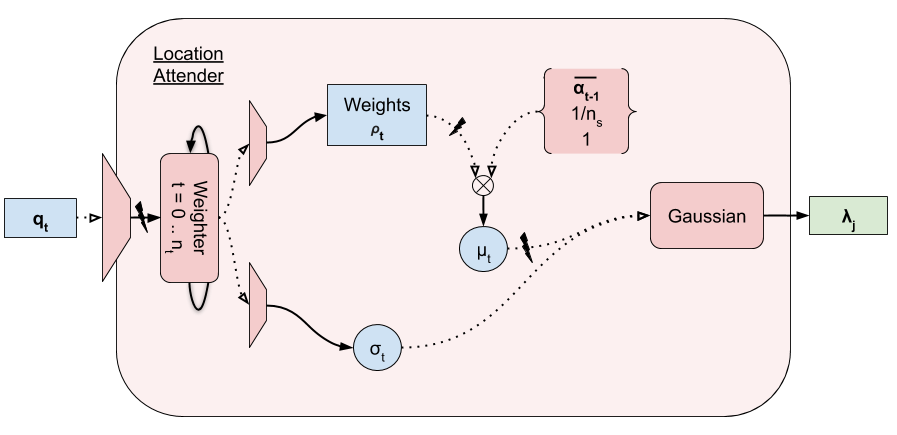}
\caption{Proposed Location Attender. Given a resized query, the Weighter outputs the standard deviation $\sigma_t$ and $\pmb{\rho}_t$ which will weight the building blocks $\mathbf{b}_t$ to compute the mean $\mu_t$. $\mu_t$ and $\sigma_t$ parametrize a Gaussian PDF used to compute the location attention $\pmb{\lambda}_t$.}
\label{fig:position}
\end{figure}

\begin{figure}[h]
\centering
\includegraphics[width=\columnwidth]{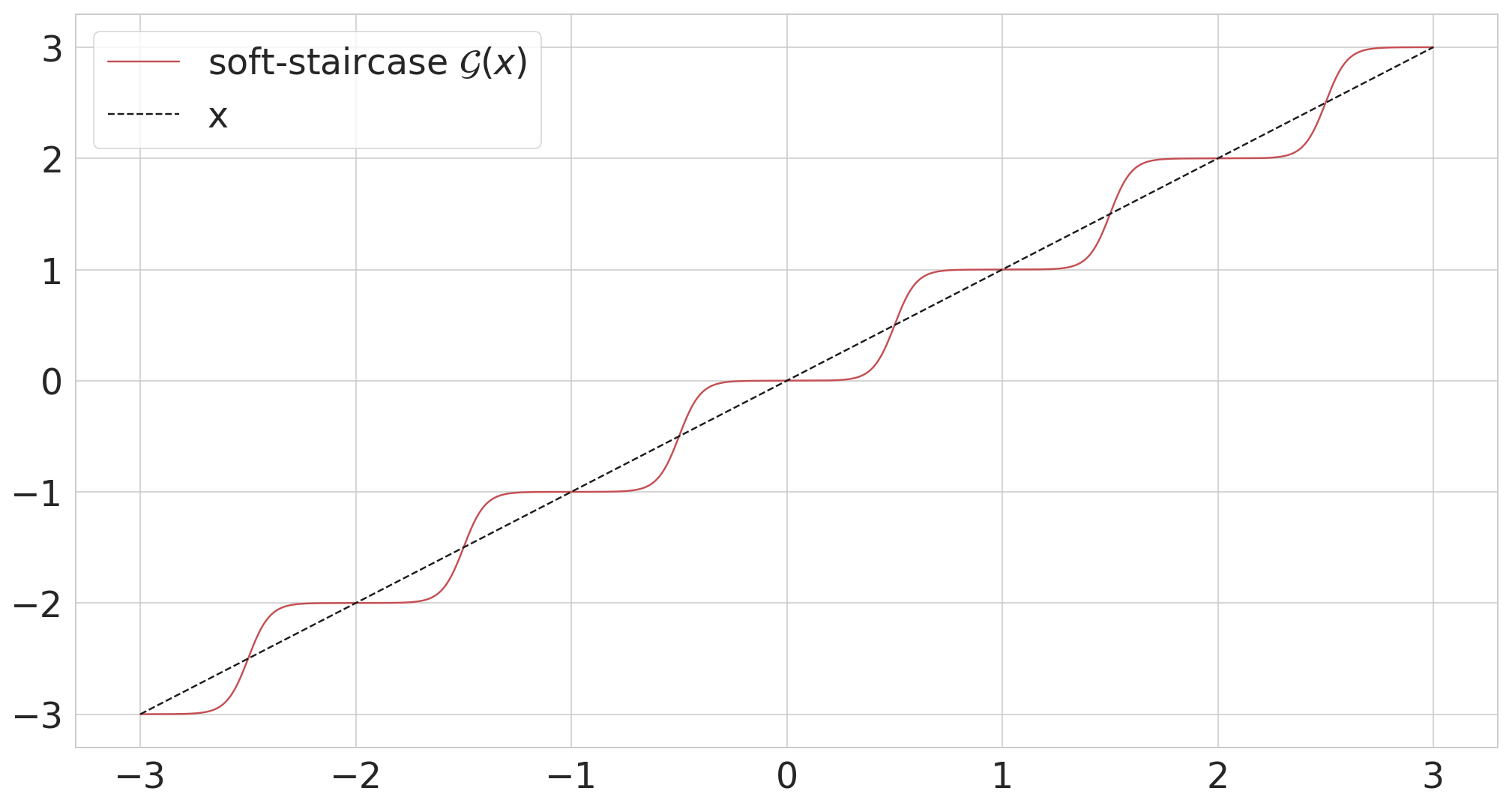}
\caption{Soft staircase activation function.}
\label{fig:soft_staircase}
\end{figure}

The weights $\pmb{\rho}_t$ are generated using a Gated Recurrent Unit (GRU) \cite{cho2014learning}. $\mu_t$ is clamped to $[0,1]$ by a linear function to yield interpretable and extrapolatable behaviour. We also force $\sigma_t > min_\sigma$ and normalize it by $n_s$ which respectively avoids division by $0$ and makes $\sigma_t$ comparable regardless of $n_s$. A graphical overview of the Location Attender can be seen in Figure~\ref{fig:position}. Formally:
\begin{align*}
\pmb{\omega}_t &\coloneqq \mathrm{GRU}\left(\mathrm{ReLU}\left(\mathrm{W}^{(resize)} \mathbf{q}_t\right)\right) \\
\sigma_t &\coloneqq \frac{\mathrm{ReLU}(\mathrm{W}^{(\sigma)}\pmb{\omega}_t) + min_\sigma}{n_s} \\
\pmb{\rho}_t &\coloneqq a(\mathrm{W}^{(\rho)}\pmb{\omega}_t) \\
\mathbf{b}_t &\coloneqq \{\bar{\pmb{\alpha}}_{t-1}; \frac{1}{n_s-1};  1\} \\
\mu_t &\coloneqq \mathrm{clamp}(\pmb{\rho}_t^T \mathbf{b}_t) \\
\lambda_{st} &\coloneqq  \frac{1}{\sqrt{2 \pi \sigma_t^2}}\exp \left(\frac{-(\frac{s}{n_s -1}-\mu_t)^2}{2 \sigma_t^2}  \right) 
\end{align*}
Where $clamp$ is a leaky clamping (2 leaky ReLUs) and $\min_\sigma = 0.27$.
$a$ is the activation function that forces each of the three dimensions of $\pmb{\rho}_t$ to take on the desired values. 
Namely a sigmoid activation for the gates, and the following ``soft-staircase'' 
\footnote{Straight-through estimators \citep{bengio2013estimating} and Gumbel-Softmax \citep{jang2016categorical,maddison2016concrete} performed slightly worst and required predefining the maximum number of steps.} 
to force the weights of the step size to be approximately integers (Figure~\ref{fig:soft_staircase}):
\[
softstair(x) \coloneqq \lfloor x \rfloor + \mathrm{sigmoid}(20 (x - 0.5 - \lfloor x \rfloor))
\]


\subsection{Mix Attender}
\label{subsection:mix_attender}

\begin{figure}[h]
\centering
\includegraphics[width=\columnwidth]{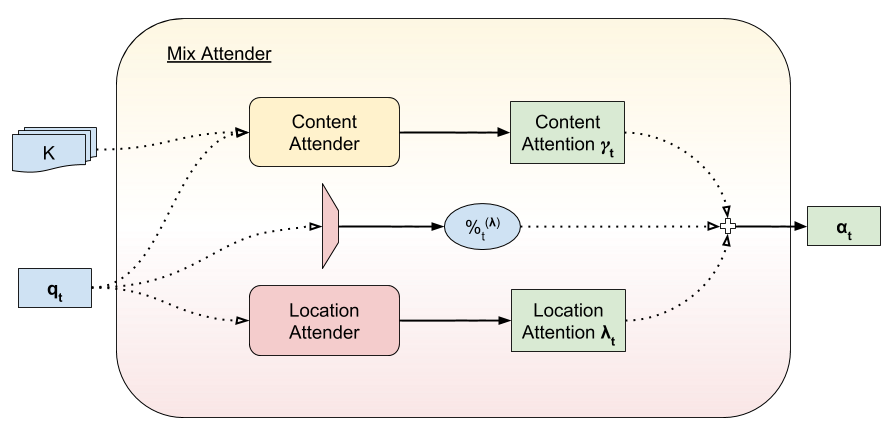}
\caption{Mix Attender. The output $\pmb \alpha_t$ is a convex combination of the content and location attention.}
\label{fig:mix}
\end{figure}

We enforce the content patterns constraint, by using a convex combination of content and location attention (Figure~\ref{fig:mix}):
\begin{align*}
\pmb{\alpha}_t &\coloneqq \%^{(\lambda)}_t \pmb\lambda_t + (1-\%^{(\lambda)}_t) \pmb\gamma_t\\
\%^{(\lambda)}_t &\coloneqq \mathrm{sigmoid}(\mathrm{W}^{(\%)}\mathbf{q}_t)
\end{align*}

\label{sec:method}

\section{Experiments}
\label{sec:experiments}

\subsection{Datasets}

The fact that humans generate and understand unbounded sentences with a finite experience is often used as proof of the \textit{principle of compositionality} \citep{sep-compositionality}.
Following this argument, methods that can extrapolate to longer sequences should exhibit some compositionality.

Based on this observation, we evaluate on a compositionality-specific artificial task, \emph{lookup tables} \citep{liska2018memorize}, but extend it to better quantify extrapolation. \footnote{ The extended datasets as well as scripts to generate them can be found at \url{https://github.com/i-machine-think/machine-tasks/tree/master/LongLookupTables}}
This task is especially interesting to us, as there is a clear notion of what a good attention pattern should look like, making it easy to qualitatively and quantitatively analyze attentive models.
It is a well-controlled task, which allows us to uncover challenges that prevent models from extrapolating on real-world data.

%



\subsubsection{Long Lookup Tables}

\begin{table}[h!]
\centering
\resizebox{\columnwidth}{!}{
	\begin{tabular}{|c|c|c|}
	\hline
	\rowcolor[HTML]{EFEFEF} 
	{\color[HTML]{000000} \textbf{Input}} & {\color[HTML]{000000} \textbf{Target}} & {\color[HTML]{000000} \textbf{Target Attention}} \\ \hline
	\rowcolor[HTML]{FFFFFF}
	{\color[HTML]{000000} 000 t1 .} & {\color[HTML]{000000} 000 110   \texttt{<eos>}} & {\color[HTML]{000000} 0 1 2} \\ \hline
	{\color[HTML]{000000} 110 t1 .} & {\color[HTML]{000000} 110 110   \texttt{<eos>}} & {\color[HTML]{000000} 0 1 2} \\ \hline
	\rowcolor[HTML]{FFFFFF}
	{\color[HTML]{000000} 110 t2 .} & {\color[HTML]{000000} 110 100  \texttt{<eos>}} & {\color[HTML]{000000} 0 1 2} \\ \hline
	\rowcolor[HTML]{FFFFFF}
	{\color[HTML]{000000} 000 t1 t1 t2 .} & {\color[HTML]{000000} 000 110 110 100 \texttt{<eos>}} & {\color[HTML]{000000} 0 1 2 3 4} \\ \hline
	\end{tabular}
}
\caption{Long lookup table examples.}
\label{table:lookup}
\end{table}

The \textit{lookup tables} task consists in sequentially applying $k$ pre-defined lookup table functions. 
The lookup tables are bijective mappings on the set of all 3-bit strings
$t_i:  \{0,1\}^3 \to \{0,1\}^3$.
For example, if $t_1(000) = 110$ and $t_2(110) = 100$ then $t_2(t_1(000))=t_2(110)=100$.
Following \citet{hupkes2018learning}, we write the operations from left to right, as well as add the inputs and temporary steps to the targets. 
E.g. the previous example corresponds to the input \texttt{000 t1 t2} and the target \texttt{000 110 100}.

General extrapolatable seq2seq models should be able to terminate by outputting an end of sentence token \texttt{<eos>}.
We thus append \texttt{<eos>} to the targets and a full stop \texttt{.} to the inputs. \footnote{This makes the task harder than the one in \citet{hupkes2018learning}, who force termination after the right amount of steps.}

At each decoding step, the target only depends on the previous output and the current lookup table.
E.g. the last decoding step of \texttt{000 t1 t2}, only depends on the previous output $110 = t_1(000)$ and the current table $t_2$.
The network thus has to learn the lookup table mappings and use the correct one at each step.
The gold standard attention, therefore, corresponds to the position of the current lookup table. Table~\ref{table:lookup} illustrates a longer example and its correct attention.

The various train and test sets are generated by composing 6 random lookup tables ${t_1, \dots, t_6}$ that have as input and output one of the $2^3=8$ possible 3-bit strings.
Specifically, we use $k=1\dots 4$ composed tables in the training set, $k=2 \dots 4$ for the interpolation test sets, and $k=5 \dots 9$ for the extrapolation test sets.

There are 5 different extrapolation test sets, depending on their additional lengths compared to the maximum training examples (\texttt{long 1}, \dots, \texttt{long 5}). 
We randomly select only 5000 possible examples for each of these test sets.

For the interpolation test sets, we select 3000 examples from all possible input-output pairs.

The training set contains all other possible input-output pairs, approximately 10000 examples.

\subsubsection{Reversed Lookup Tables}

To test whether the attention can generate more complex patterns (investigating the Positional Patterns Constraint), we also introduce a dataset which reverses the order of the inputs in the previous dataset. 
E.g. the last example in Table~\ref{table:lookup}, would be written as \texttt{t2 t1 t1 000 .}, the target would not change, and the attention pattern should be \texttt{3 2 1 0 4} (attend to \texttt{.} when outputting \texttt{<eos>}). 
Although the change seems minor, we hypothesize that such a setting will be much more complicated as the attention pattern is not monotonic and does not follow the encoding nor the decoding steps.
Indeed, in the previous task, the model 
only needs to learn to match the $i^{th}$ decoding step with the $i^{th}$ encoding step.

\subsubsection{Lookup Tables with Noisy Start}

\begin{table}[h!]
\centering
\resizebox{\columnwidth}{!}{
\begin{tabular}{|c|c|c|}
\hline
\rowcolor[HTML]{EFEFEF} 
{\color[HTML]{000000} \textbf{Input}} & {\color[HTML]{000000} \textbf{Target}} & {\color[HTML]{000000} \textbf{Target Attention}} \\ \hline
\rowcolor[HTML]{FFFFFF} 
{\color[HTML]{000000} 000 t2 ! t1 .} & {\color[HTML]{000000} 000 110 \texttt{<eos>}} & {\color[HTML]{000000} 0 2 3 4} \\ \hline
\rowcolor[HTML]{FFFFFF} 
{\color[HTML]{000000} 110 t5 t3 t1 ! t1 .} & {\color[HTML]{000000} 110 110 \texttt{<eos>}} & {\color[HTML]{000000} 0 4 5 6} \\ \hline
\rowcolor[HTML]{FFFFFF} 
{\color[HTML]{000000} 110 ! t2 .} & {\color[HTML]{000000} 110 100 \texttt{<eos>}} & {\color[HTML]{000000} 0 2 3} \\ \hline
\rowcolor[HTML]{FFFFFF} 
{\color[HTML]{000000} 000 t6 t3 ! t1 t1 t2 .} & {\color[HTML]{000000} 000 110 110 100 \texttt{<eos>}} & {\color[HTML]{000000} 0 3 4 5 6 7} \\ \hline
\end{tabular}
}
\caption{Lookup tables with noisy start examples}
\label{table:noisy}
\end{table}

Finally, we introduce another variant that also requires content attention (investigating the Content Patterns Constraint). 
To do so, we augment each training example with a start token ``\texttt{!}'' between the input and the tables in the source sequence. We then add $m \sim \mathcal{U}\{0,10\}$ tables $t_i$ before the start token. The target outputs were not modified and are thus independent of the added tables. Solving this task requires to first attend to the input, then to the token which follows ``\texttt{!}'' (content attention) and finally proceed with incremental location attention.  Examples of the training data are given in Table~\ref{table:noisy}.

\subsection{Metrics}
\label{ssec:metrics}

The main metric is \textit{sequence accuracy} (\textit{seqAcc}), which corresponds to the accuracy of predicting the entire sequence correctly (including its length). To get insights about how the model works, we will also use two other losses. 

\textit{Sequence Accuracy Before Eos} (\textit{seqAccBE}), which only evaluates the accuracy of the sub-sequence before the model generated a \texttt{<eos>}.

\textit{Attention Loss} (\textit{attnLoss}), which quantifies the quality of the attention pattern before \texttt{<eos>}. It is computed as the mean squared error between the predicted and gold standard attention. \footnote{The loss is overly simplistic as it is symmetric around $\bar{\pmb\alpha}_t$ even though errors in the temporal direction are less serious as the embeddings contain past information.} The attention loss gives an indication of how far the model is to the ideal attention patterns required to solve the sequence. 

\subsection{Architecture and Baselines}

Concerning baselines, we use three content attention: additive, multiplicative, scaled dot product (Eq.\ref{eq:content}).
We also have two mixed content-location attention baselines: Transformer and TransformerXL (Eq.\ref{eq:location}).


To focus on the attention mechanisms, our model and the baselines all use a smaller version of the best performing recurrent seq2seq architecture on the \textit{lookup table} task \cite{hupkes2018learning}. 
The model has never been modified during our experimentation and is schematized in Figure~\ref{fig:attender_seq2seq}.
The embeddings are of dimension $64$, the recurrent network is a GRU \cite{cho2014learning} with a hidden size of $128$, $50\%$ dropout \cite{srivastava2014dropout} is applied on the encoder-decoder bottleneck, and a residual connection is used between the inputs (embeddings) and outputs of the encoder.
Training consists of 50 epochs with the Adam \cite{kingma2014adam} optimizer. 

\section{Results}
\label{ref:results}
\subsection{Interpolation}

For sanity check, we tested all the baselines and our models (with and without attention mix) on the interpolation setting of the three tasks.
Our models and the best baseline (transformer attention) achieved $100\%$ sequence accuracy (\textit{seqAcc}).

\subsection{Extrapolation Constraint}

\begin{figure}[h]
\centering
\includegraphics[width=\columnwidth]{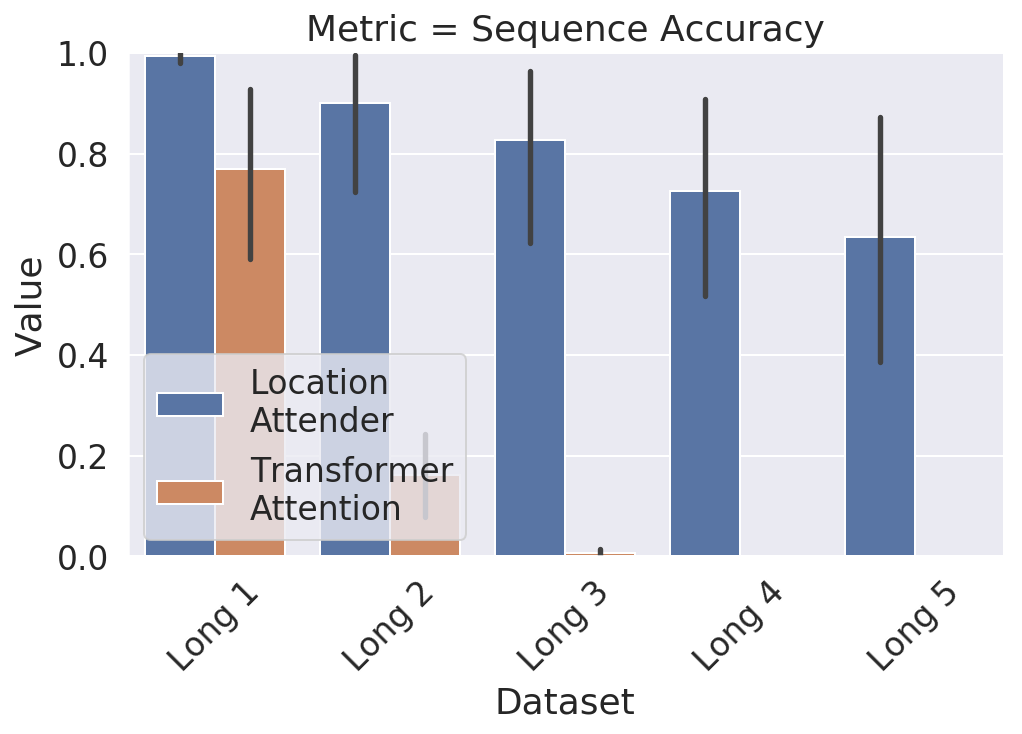}
\caption{\textit{SeqAcc} for the Location Attender and best baseline on the Long Lookup Tables task ($10$ runs).}
\label{fig:res_loc}
\end{figure}

The major desired property of our model is to be able to extrapolate. 
We tested the extrapolation capacity
of our location attender by evaluating its \textit{seqAcc} on the long lookup table extrapolation test sets.
Figure~\ref{fig:res_loc} shows the \textit{seqAcc} of the location attender against the strongest baseline (transformer attention). 

As hypothesized, the transformer attention has some extrapolation capacity, but our location attender substantially outperforms it in this simple task. 
Importantly, the loss in performance in the extrapolation setting for the best baseline is abrupt and goes from $100\%$ to $0\%$ by adding only three tokens to the inputs. 
This suggests that commonly used models are brittle and cannot even extrapolate by a small amount.

Although the previous results are encouraging, we would like to understand what is holding back our model from perfectly extrapolating (Figure~\ref{fig:res_loc}).

\begin{figure}[h]
\centering
\includegraphics[width=\columnwidth]{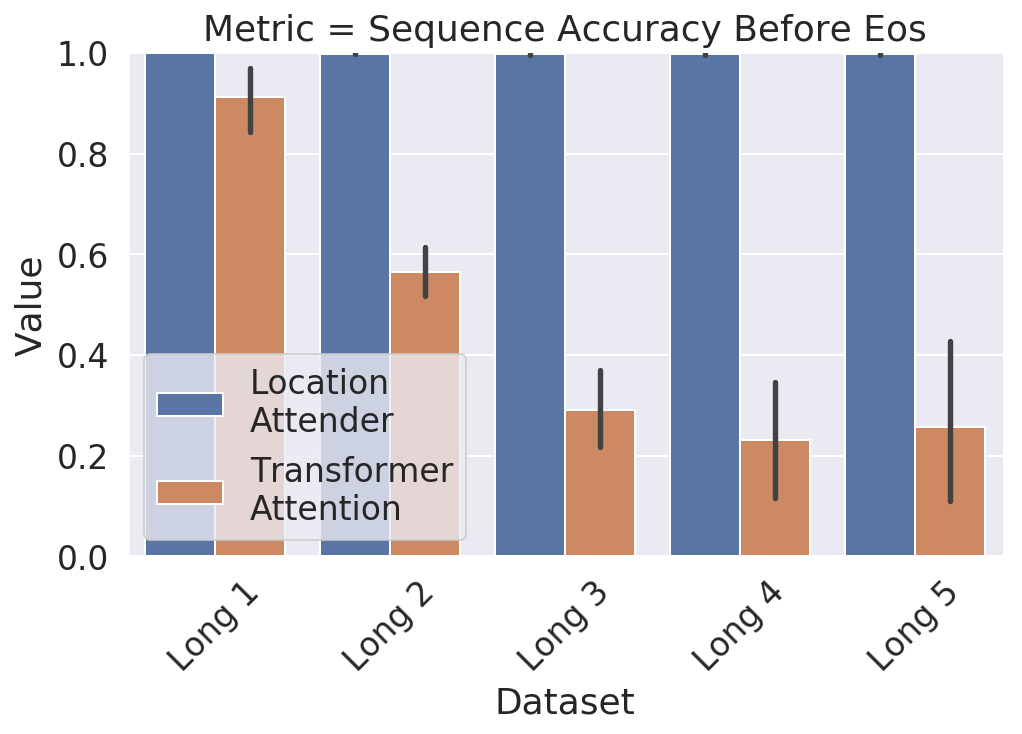}
\caption{\textit{SeqAccBE} for the Location Attender and best baseline on the Long Lookup Tables task ($10$ runs).}
\label{fig:res_loc_before}
\end{figure}

To do so, we computed the sequence accuracy before \texttt{<eos>} (\textit{SeqAccBE}).
Figure~\ref{fig:res_loc_before} shows that the model outputs are always correct but that it often terminates decoding too soon, which we will refer to as \textbf{the \texttt{<eos>} problem}. 
This suggests that the decoder keeps an internal ``counter'' to increase the probability of outputting \texttt{<eos>} when the decoding step is greater than the ones seen at training time. 
The model learns this heuristic, which is always correct during training time and can be thought of as a metric hacking. 
Importantly, it is not a ``hard'' boundary: the model is often able to extrapolate a couple of steps but usually stops before the correct number of steps.

\subsection{Positional and Content Patterns Constraint}

Having shown that our model can extrapolate well on a simple task, we would like to investigate whether it can do so for tasks that require more complicated attention patterns such as the reversed and noisy task. 

Although the Mix Attender, outperformed all baselines on both tasks, it was not able to get more than $40\%$ and $5\%$ sequence accuracy for \texttt{long 1} and \texttt{long 2} respectively.

\begin{figure}[h!]
\centering
\centerline{\includegraphics[width=\columnwidth]{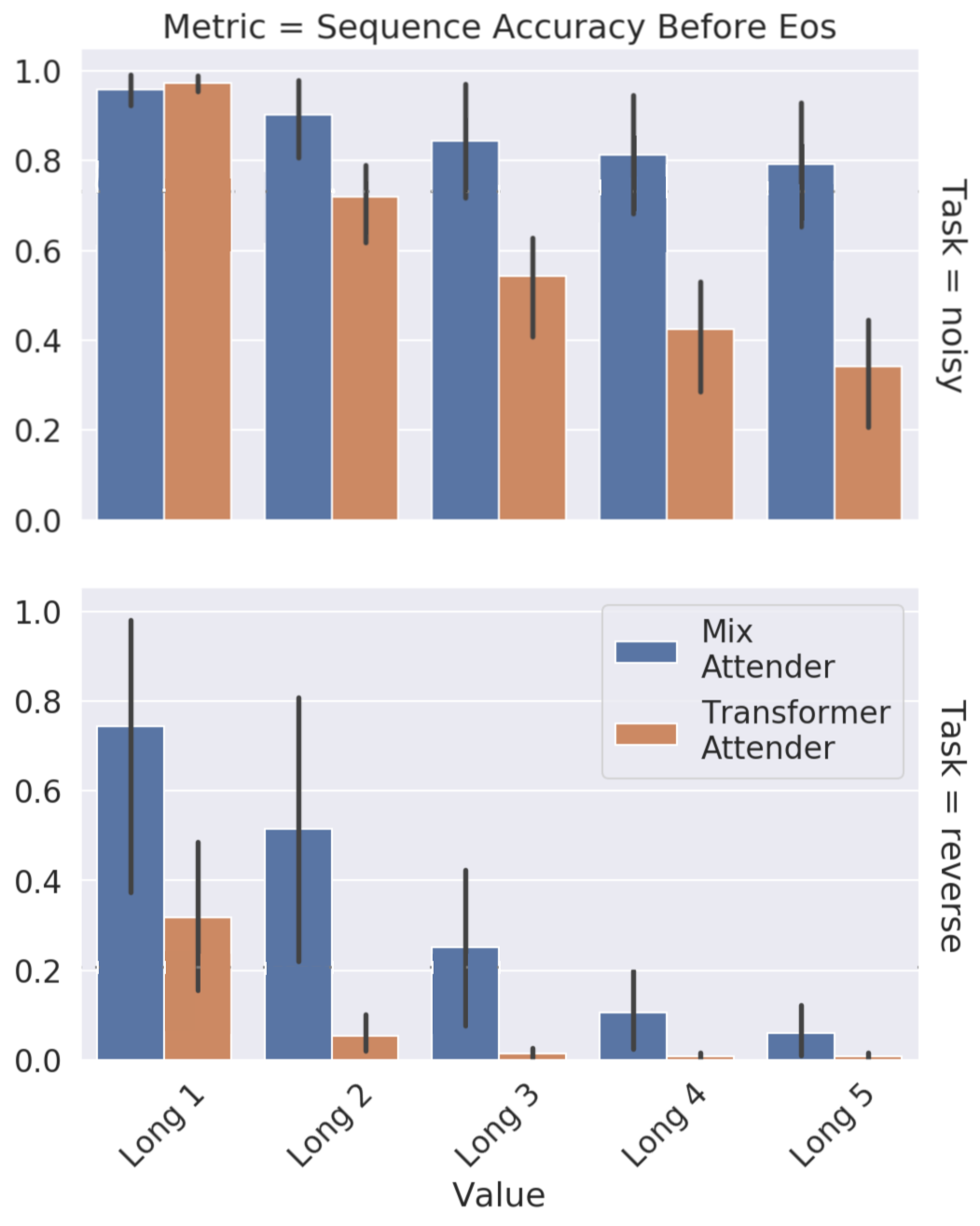}}
\caption{\textit{SeqAccBE} ($5$ runs) for the Mix Attender and best baseline on the reversed lookup tables (\textit{reverse}) and lookup tables with noisy start (\textit{noisy}).}
\label{fig:res_mix}
\end{figure}

Figure~\ref{fig:res_mix} shows that when considering \textit{seqAccBE}, the Mix Attender is able to extrapolate well in the noisy setting and a little in the reverse setting. 
This suggests that it is not able to extrapolate well when considering sequence accuracy because it strongly suffers from the \texttt{<eos>} problem.
This is a recurrent problem in our experiments and is more likely to happen in harder tasks and larger models.

\subsection{Attention Pattern}

As previously discussed, variants of the lookup table task are especially interesting as we know the gold standard attention pattern. 
This enables evaluation of attention patterns through the MSE attention loss (\textit{attnLoss}). 

\begin{table}[h]
\centering
\resizebox{\columnwidth}{!}{
\begin{tabular}{|
>{\columncolor[HTML]{EFEFEF}}c |
>{\columncolor[HTML]{FFFFFF}}c |
>{\columncolor[HTML]{FFFFFF}}c |
>{\columncolor[HTML]{FFFFFF}}c |
>{\columncolor[HTML]{FFFFFF}}c |
>{\columncolor[HTML]{FFFFFF}}c |
>{\columncolor[HTML]{FFFFFF}}c |}
\hline
{\color[HTML]{000000} \textbf{Attention}} & \cellcolor[HTML]{EFEFEF}{\color[HTML]{000000} \textbf{Interp.}} & \cellcolor[HTML]{EFEFEF}{\color[HTML]{000000} \textbf{Long 1}} & \cellcolor[HTML]{EFEFEF}{\color[HTML]{000000} \textbf{Long 2}} & \cellcolor[HTML]{EFEFEF}{\color[HTML]{000000} \textbf{Long 3}} &
\cellcolor[HTML]{EFEFEF}{\color[HTML]{000000} \textbf{Long 4}} &
\cellcolor[HTML]{EFEFEF}{\color[HTML]{000000} \textbf{Long 5}} \\ \hline
{\color[HTML]{000000} \textbf{Scaled Dot}} & {\color[HTML]{000000} 5.3} & {\color[HTML]{000000} 6.3} & {\color[HTML]{000000} 8.1} & {\color[HTML]{000000} 10.2} &
{\color[HTML]{000000} 12.6} &
{\color[HTML]{000000} 15.4} \\ \hline
{\color[HTML]{000000} \textbf{Multiplicative}} & {\color[HTML]{000000} 3.1} &
{\color[HTML]{000000} 4.6} & {\color[HTML]{000000} 6.3} & {\color[HTML]{000000} 7.9} & {\color[HTML]{000000} 9.9} & {\color[HTML]{000000} 12.4} \\ \hline
{\color[HTML]{000000} \textbf{Additive}} & {\color[HTML]{000000} 3.1} & {\color[HTML]{000000} 8.4} & {\color[HTML]{000000} 15.6} & {\color[HTML]{000000} 22.2} &
{\color[HTML]{000000} 28.7} &
{\color[HTML]{000000} 34.8} \\ \hline
{\color[HTML]{000000} \textbf{Transformer}} & {\color[HTML]{000000} 2.8} & {\color[HTML]{000000} 3.5} & {\color[HTML]{000000} 6.1} & {\color[HTML]{000000} 9.1} &
{\color[HTML]{000000} 11.7} &
{\color[HTML]{000000} 13.9} \\ \hline
{\color[HTML]{000000} \textbf{TransformerXL}} & {\color[HTML]{000000} 3.0} & {\color[HTML]{000000} 3.9} & {\color[HTML]{000000} 5.3} & {\color[HTML]{000000} 7.1} &
{\color[HTML]{000000} 9.1} &
{\color[HTML]{000000} 11.4} \\ \hline
{\color[HTML]{000000} 
\textbf{Mix Attention}} & {\color[HTML]{000000} \textbf{2.1}} & {\color[HTML]{000000} \textbf{2.2}} & {\color[HTML]{000000} \textbf{2.9}} & {\color[HTML]{000000} \textbf{4.1}} &
{\color[HTML]{000000} \textbf{5.3}} &
{\color[HTML]{000000} \textbf{6.7}} \\ \hline
\end{tabular}
}
\caption{\textit{AttnLoss} for various attention models averaged over the three datasets and 5 runs.}
\label{tab:mean_attn}
\end{table}

Table~\ref{tab:mean_attn} shows the attention loss averaged over the three tasks.
Although not perfect, the Mix Attender performs on average the best across all settings. \footnote{Some baselines outperformed it in the interpolation settings of specific tasks. Namely, the additive attention in the reversed task and transformer in the noisy task.}
Crucially, it performs similarly in an interpolation setting and simple extrapolation setting (\texttt{long 1}), while all other baselines perform significantly worse after adding a single token.
Even in \texttt{long 2}, it is competitive with all other attention mechanisms in their interpolation domain.
This indicates that the model is indeed able to extrapolate by being more precise with its attention pattern.

\subsection{Qualitative Analysis}

\begin{figure*}[h]
\centering
\centerline{\includegraphics[width=1.8\columnwidth]{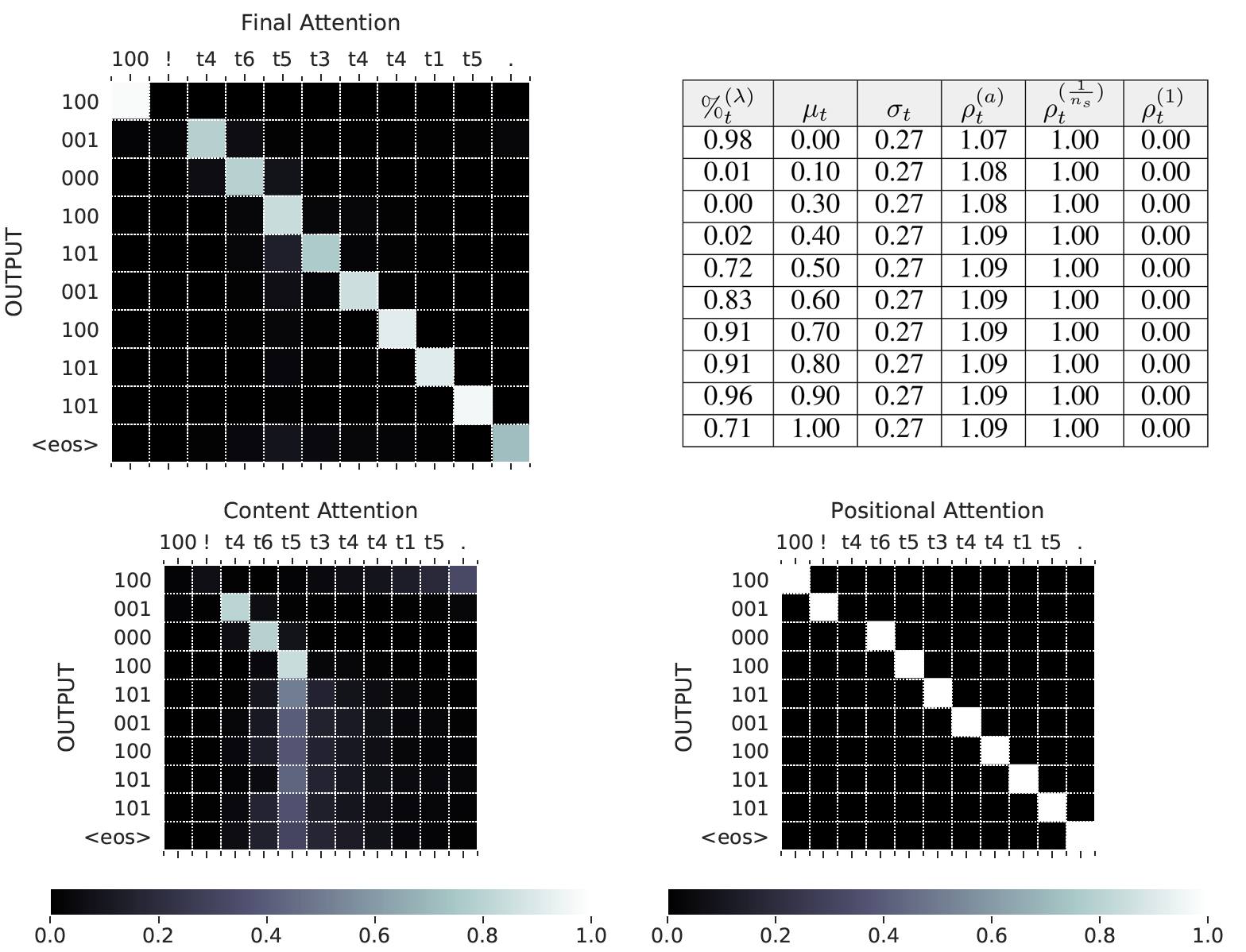}}
\caption{Example output by the attention-mixer for the lookup tables with noisy start task (\texttt{Long 4} test set).}
\label{fig:example_output}
\end{figure*}

In addition to enabling extrapolation, the temporary variables such as the weight given to each building block are very helpful for debugging the model and improving interpretability.

Figure~\ref{fig:example_output} shows the output of a Mix Attender for the \emph{lookup tables with noisy start} task.
The input was sampled from the \texttt{Long 4} test set. 
The top-left image shows the final attention.
The top-right table shows the value of some interpretable variables at every decoding step. 
The bottom images correspond to the content and location attention. 

The first decoding step uses location attention to attend to the first input. 
For the next three steps, the model outputs a mixing weight $\%^{(\lambda)} \approx 0$ to focus on content attention. 
The content attention successfully finds the first non-noisy table (after \texttt{!}). \footnote{A single step of content attention should be sufficient, but the model seems to consistently use three steps.}
It then goes back to using the location attention with $\rho^{(\alpha)}=1$ and $\rho^{(1/n)}=1$ to generate a diagonal attention. 
Finally, it predicts \texttt{<eos>} when attending to the end of the input \enquote{\texttt{.}}.

At each step, $\sigma = min_\sigma$ as it does not need to attend to neighboring words for this task.
$\%^{(\lambda)}$ is never exactly 0 or 1, such that the model can easily learn to switch between content and location attention as it does not collapse to using a single form of attention.


\section{Discussion}



In this paper, we focused on one type of extrapolation, which is especially important in NLP: generalization to longer sequences.
We propose a new location-based attention, and show that it can extrapolate better than previous models while learning various attention patterns.

Despite promising initial results, our model is still unable to extrapolate perfectly for harder tasks.
By analyzing its behavior, we uncovered an interesting heuristic used by seq2seq models, namely that they keep track of a decoding ``counter'' to know when to output the \texttt{<eos>} token. 
This is a bottleneck for extrapolation, suggesting that removing this heuristic is key to reaching perfect extrapolation and should be investigated in future work.

Once the \texttt{<eos>} problem is solved, we could test the model on real-world datasets.
It would also be interesting to test such attention mechanisms in self-attentive seq2seq models without recurrence. 
Finally, as the location attender is not model dependent, it could be pretrained on complex location patterns and incorporated as a plug-and-play module to get extrapolatable position attention.
 



Taking a step back, we have shown that current deep learning models with common attention mechanisms are unable to extrapolate well on seemingly straightforward tasks.
This tends to be overlooked by the field due to standard benchmarks that can be solved using only interpolation.
We hope that this paper acts as a reminder that extrapolation is a hard setting that has not been much investigated by the machine learning community. 
As current methods that memorize and learn superficial cues are unable to extrapolate while humans are, we believe that such a setting might help (and force) the field to come up with more human-like computational models that are capable of abstract reasoning.

\section*{Acknowledgments}

Dieuwke Hupkes is funded by the Netherlands Organization
for Scientific Research (NWO), through a Gravitation Grant 024.001.006 to the Language in Interaction Consortium. Elia Bruni is funded by the European Union’s Horizon 2020 research and innovation program under the Marie Sklodowska-Curie
grant agreement No 790369 (MAGIC).

\bibliographystyle{acl_natbib}
\bibliography{acl2020}

\appendix

\end{document}